\documentclass{article}

\usepackage{graphicx}
\usepackage{amssymb}
\usepackage{latexsym}
\usepackage{graphicx}
\usepackage{epstopdf}
\usepackage{caption}
\usepackage{url}
\usepackage{xcolor}
\usepackage{amsmath}
\usepackage{color, colortbl}
\usepackage{enumitem}

 \usepackage{here}
\usepackage{lineno}

\usepackage{PRIMEarxiv}

\usepackage[utf8]{inputenc} 
\usepackage[T1]{fontenc}    
\usepackage{hyperref}       
\usepackage{url}            
\usepackage{booktabs}       
\usepackage{amsfonts}       
\usepackage{nicefrac}       
\usepackage{microtype}      
\usepackage{lipsum}
\usepackage{fancyhdr}       
\usepackage{graphicx}       
\graphicspath{{media/}}     
\usepackage{makecell}
\title{Analyzing the Performance of ChatGPT in Cardiology and Vascular Pathologies}

\author{
  Walid Hariri \\
  Labged Laboratory, Computer Science department\\ Badji Mokhtar Annaba University  \\
  Annaba, Algeria\\
  hariri@labged.net}

\begin{document}
\maketitle

\begin{abstract}
The article aims to analyze the performance of ChatGPT, a large language model developed by OpenAI, in the context of cardiology and vascular pathologies. The study evaluated the accuracy of ChatGPT in answering challenging multiple-choice questions (QCM) using a dataset of 190 questions from the Siamois-QCM platform. The goal was to assess ChatGPT potential as a valuable tool in medical education compared to two well-ranked students of medicine. The results showed that ChatGPT outperformed the students, scoring 175 out of 190 correct answers with a percentage of 92.10\%, while the two students achieved scores of 163 and 159 with percentages of 85.78\% and 82.63\%, respectively. These results showcase how ChatGPT has the potential to be highly effective in the fields of cardiology and vascular pathologies by providing accurate answers to relevant questions. 

\end{abstract}


\section{Introduction}
\label{sec:intro}
ChatGPT, a large language model developed by OpenAI, has gained significant attention for its potential in various domains, including the field of medicine \cite{openAI}. As a language model trained on vast amounts of text data, ChatGPT has demonstrated the ability to generate coherent and contextually appropriate responses to a wide range of queries \cite{liebrenz2023generating,hariri2023unlocking}. In the medical domain, ChatGPT has shown promise as a tool for medical education and exam preparation, particularly in assisting students in their residency exams \cite{sallam2023chatgpt}.

The potential in medical education and exams is particularly relevant and very challenging in the context of cardiology and vascular pathologies \cite{zaabi2023review,hariri2024sentiment}. These specialized areas require a deep understanding of complex medical concepts and the ability to accurately answer questions and provide relevant explanations. Therefore, the dataset described below will be a very good challenge for both students and ChatGPT. 

\section{Dataset}
\label{sec:data}
The performance of ChatGPT and two medical students in cardiology and vascular pathologies was evaluated using a dataset of medical exams from the Siamois-QCM platform. \cite{siamois}, which provides multiple-choice questions (QCM) in French language to assist students in their residency exam preparation. This platform contains more than 50,000 medical, pharmacy, and dental students to prepare for their residency exams and competitions. The platform provides the possibility to choose the materials, and also specific lessons. Each lesson has a different number of questions. The students have chosen both the material (cardiology and vascular pathologies) and the seven lessons to make the comparison to ChatGPT more challenging.

\begin{figure}[h]
\centering
\includegraphics[width=0.65\textwidth]{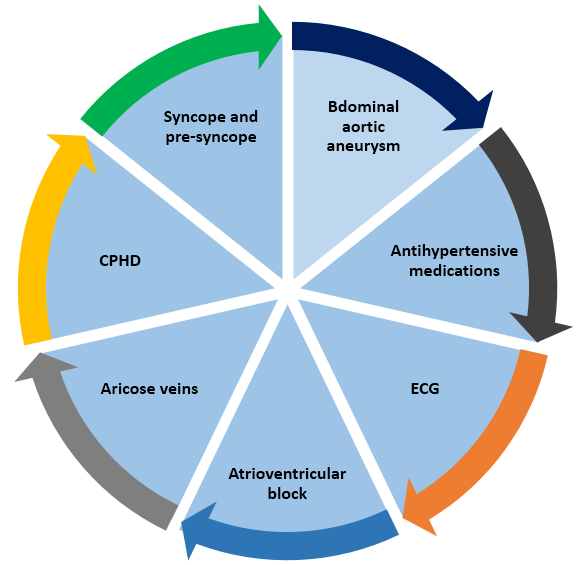}
\caption{The seven lessons related to cardiology and vascular pathologies material.}
\label{fig:seven}
\end{figure}

\begin{figure}[h]
\centering
\includegraphics[width=0.85\textwidth]{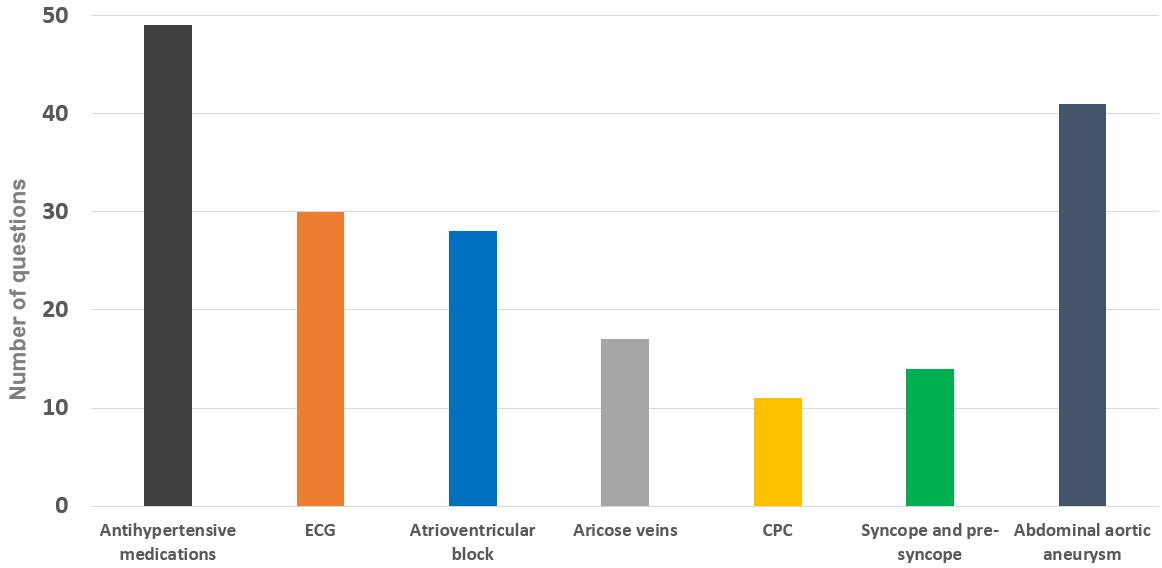}
\caption{The number of questions per lesson from Siamois-QCM platform.}
\label{fig:seven2}
\end{figure}

We will be analyzing a total of 190 questions related to the lessons presented in Figure \ref{fig:seven}. The distribution of the 190 questions on the lessons is shown in Figure \ref{fig:seven2}

\section{Methodology}
\label{sec:meth}
This article aims to analyze the performance of ChatGPT in the material of "cardiology and vascular pathologies" by utilizing a dataset of questions from Siamois-QCM platform and assessing its accuracy in answering the questions. The selected questions belong to the $6^{th}$ year Medicine faculty program of Algiers, Algeria, known for being a challenging exam. We then compare the results of ChatGPT with the performance of two well-ranked students of medicine who are currently studying in the same program.


This study specifically focuses on seven lessons within the material of "cardiology and vascular pathologies". Below, we provide a brief overview of lesson and the related disease:
\begin{itemize}
   
\item \textbf{Abdominal aortic aneurysm:} A bulge in the lower part of the aorta, which can be life-threatening if it ruptures.
\item \textbf{Antihypertensive medications:} Drugs used to lower high blood pressure and prevent associated complications such as stroke and heart attack.
\item \textbf{Normal and pathological electrocardiograms (ECG):}  A test that records the electrical activity of the heart to diagnose heart problems, including abnormal rhythms and damage.
\item \textbf{Atrioventricular block:} A condition where the electrical signals between the upper and lower chambers of the heart are disrupted, resulting in a slower heart rate or irregular heartbeat.
\item \textbf{Varicose veins:} Twisted, enlarged veins, usually in the legs, that can cause pain and discomfort.
\item \textbf{Chronic pulmonary heart disease (CPC):} A condition where the lungs and heart are unable to function properly due to long-term lung disease.
\item \textbf{Syncope and pre-syncope:} Fainting or feeling like one might faint, often caused by a drop in blood pressure, lack of oxygen to the brain, or other underlying medical conditions.
\end{itemize}

\section{Results}
\begin{figure}[h]
\centering
\includegraphics[width=0.63\textwidth]{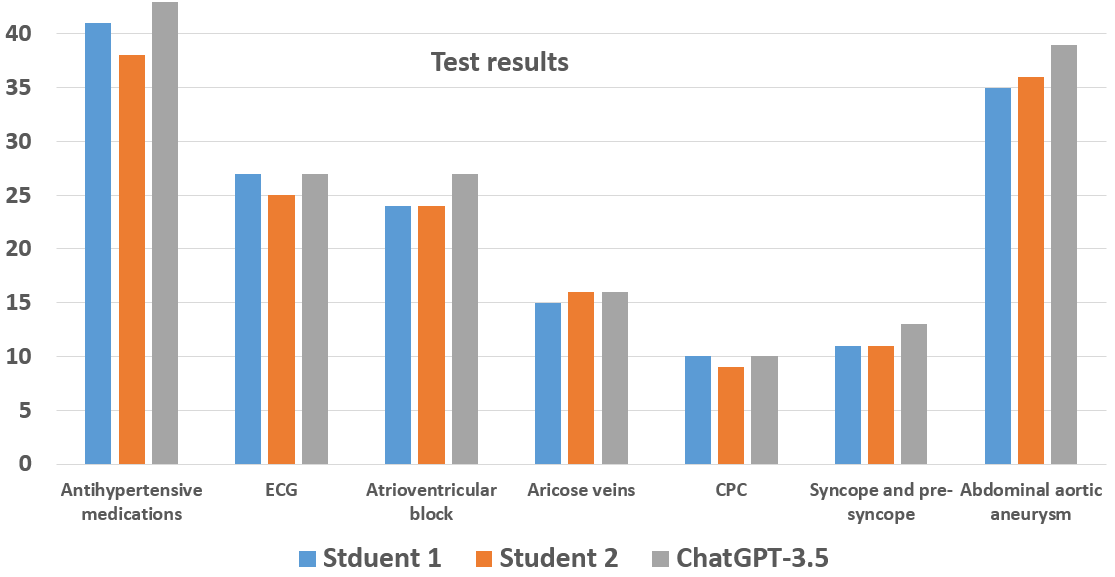}
\caption{Test results of the two students and ChatGPT per lesson.}
\label{fig:bars2}
\end{figure}

\begin{figure}[h]
\centering
\includegraphics[width=0.99\textwidth]{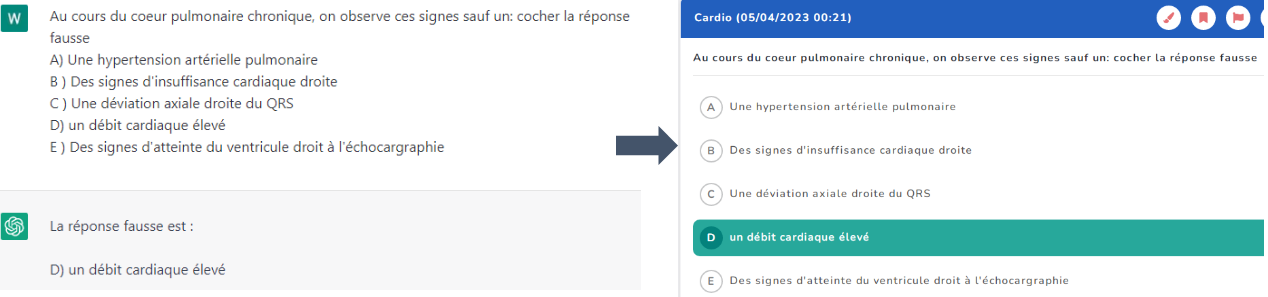}
\caption{Example of a correct answer from ChatGPT on Siamois-QCM platform.}
\label{fig:resq1}
\end{figure}

\begin{figure}[h]
\centering
\includegraphics[width=0.99\textwidth]{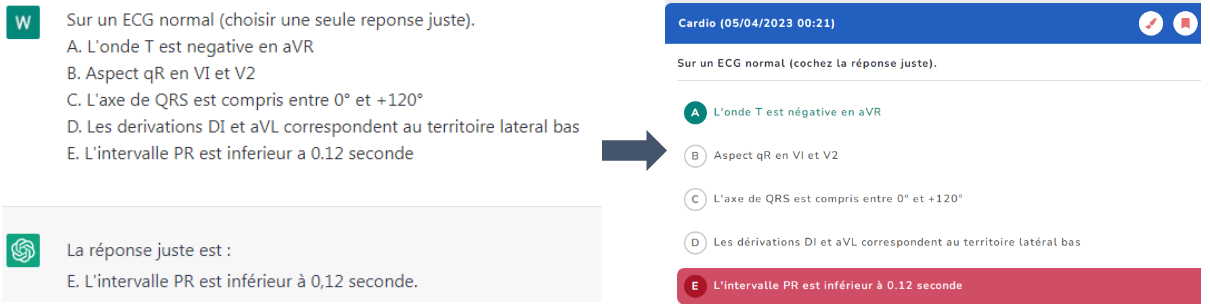}
\caption{Example of a wrong answer from ChatGPT on Siamois-QCM platform.}
\label{fig:resq2}
\end{figure}

The findings of this study highlight the potential of ChatGPT as a valuable tool in medical education within the material of "cardiology and vascular pathologies. Table \ref{tab:results} shows that ChatGPT outperformed the scores of the two well-ranked students by achieving 175 correct answers out of 190 questions, with a percentage of 92.10\%. On the other hand, the two students achieved scores of 163 and 159, with a percentage of 85.78\% and 82.63\% respectively. Figure \ref{fig:bars2} summarizes the test results per lesson. Figures \ref{fig:resq1} and \ref{fig:resq2} depict a correct and an incorrect answer, respectively, from ChatGPT on the Siamois-QCM platform. Therefore, questions that contain numerical values with different units may prove more challenging for ChatGPT, potentially resulting in incorrect answers.

\begin{table*}[h]
\footnotesize
\centering
\caption{Number of correct answers, for 190 questions answered by ChatGPT and two students.}
\begin{tabular}{|l|c|c|c|c|}\hline 
\textbf{Lesson} & \textbf{\makecell{Number of \\ questions}}  & \textbf{Student 1}  &\textbf{Student 2} & \textbf{ChatGPT} \\ \hline   
Abdominal aortic aneurysm     & 41  & 35   & 36 & 39   \\ \hline 
Antihypertensive medications  &49   & 41   & 38 &  43 \\ \hline 
ECG                           & 30 & 27    & 25 &  27\\ \hline 
Atrioventricular block        & 28  & 24   & 24  & 27  \\ \hline 
Varicose veins                & 17 & 15    & 16 &  16 \\ \hline 
 CPC                          & 11 & 10    &9    &  10 \\ \hline 
Syncope and pre-syncope       & 14  & 11   & 11  &  13  \\ \hline 
\textbf{Total}                & 190 &163   & 159  & \textbf{175}  \\ \hline 
\textbf{Percentage}          & 100\% &85.78\%   & 83.68\%  & \textbf{92.10\%}  \\ \hline 
\end{tabular}
\label{tab:results}
\end{table*}

\section{Conclusion}
This paper demonstrates the high potential of ChatGPT in the field of cardiology and vascular pathologies by providing answers to related questions from Siamois-QCM platform. Although ChatGPT has outperformed the score of the 2 well-ranked students with a 6\% advantage, it is necessary to provide further refinement and improvement in its performance for specific medical domains. Further research and development are necessary to enhance ChatGPT capabilities for assisting students in residency exam preparation and supporting medical education in this specialized field. This analyzing study will be expanded to include more materials to further evaluate ChatGPT's performance in the medical field.
\bibliographystyle{unsrt}  
\bibliography{references}

\end{document}